\documentclass{IOS-Book-Article}
\usepackage{soul}\setuldepth{article}

\usepackage{multirow}
\usepackage{mathptmx}
\usepackage{comment}
\usepackage{graphicx}
\usepackage{todonotes}
\usepackage{subcaption}
\usepackage{algorithm}
\usepackage{algpseudocode}
\usepackage{amsmath}
\usepackage{listings}

\newcommand{\furl}[2]{\footnote{#1: \url{#2}}} 

\def\hb{\hbox to 11.5 cm{}}

\begin{document}

\pagestyle{headings}
\def\thepage{}
\begin{frontmatter}              

\title{From Historical Tabular Image to Knowledge Graphs: A Modular, Provenance-Aware  Pipeline}

\markboth{}{May 2026\hb}

\author[A]{\fnms{Sarah Binta Alam} \snm{Shoilee}\orcid{0000-0001-9458-8105}%
\thanks{Corresponding Author: Sarah Binta Alam Shoilee, s.b.a.shoilee@vu.nl}},
\author[A]{\fnms{Victor} \snm{de Boer}\orcid{0000-0001-9079-039X}}
\author[A]{\fnms{Jacco} \snm{van Ossenbruggen}\orcid{0000-0002-7748-4715}}
and
\author[A]{\fnms{Susan} \snm{Legêne}\orcid{0000-0002-2826-9541}}

\runningauthor{S.B.A. Shoilee et al.}
\address[A]{Vrije Universiteit Amsterdam, Amsterdam, the Netherlands}


\begin{abstract}
Handwritten archival tables contain rich historical information, yet transforming them into structured representations, such as knowledge graphs (KGs), requires integrating table structure recognition, handwriting recognition, and semantic interpretation—a complex multimodal process. End-to-end AI implementations can obscure these steps, resulting in opaque algorithmic operations that hinder human oversight, critical assessment, and trust. 
To address this, we present a modular, provenance-aware pipeline to convert handwritten tabular images into KGs supporting human–AI collaboration. The pipeline decomposes the workflow into three separate stages—table reconstruction, information extraction, and KG construction—while exposing intermediate representations for inspection, evaluation, and correction. A key contribution of our approach is the systematic integration of data provenance at every stage, ensuring that all extracted entities and literals remain traceable to their visual and textual origins. 
The proposed approach is validated through a number of demonstrations on real-world archival material concerning military careers. The demonstration across three different table reconstruction methods highlights the importance of modularisation. Additionally, the use-case demonstration of the automatic extraction of military career events with provenance preservation illustrates the pipeline's ability to support controlled human intervention, which remains necessary to ensure data authenticity and facilitate correction of information.
By coupling modularity with data provenance, our work advances transparent and collaboratively controllable image-to-KG pipelines for complex historical data.


\end{abstract}

\end{frontmatter}
\markboth{May 2026\hb}{May 2026\hb}
\clearpage
\section{Introduction}




Knowledge graphs (KG) have become a central infrastructure for publishing and integrating data across domains \cite{hogan2021knowledge}, including cultural heritage and archival institutions, enabling structured representation of institutional knowledge, semantic linking and inter-collection inter-operation \cite{deBoer2025}. Using different semantic web technologies, heterogeneous datasets published as knowledge graphs can be interconnected, queried, and analysed in ways that support both scholarly research and public accessibility. Recent advances in artificial intelligence further strengthen this ecosystem by automating data extraction and enrichment processes \cite{kraevandluchev2025analyzing, zhang2023towards}. However, a large portion of historically rich information still exists only as scanned documents (images) and therefore remains inaccessible to machine-interpretable workflows. Many historical administrative records—such as registers, census tables, and military records—were originally recorded in tabular form \cite{merono-penuela2020ontologies}. Transforming these handwritten tables into structured knowledge for large-scale analysis requires solving multiple interdependent problems simultaneously: table structure recognition, handwritten text recognition, and knowledge graph construction \cite{merono-penuela2020ontologies,tuominen2017bioCRM}.


KG construction is the process of transforming structured or unstructured data into a structured knowledge graph by creating entities and linking them through semantic relationships, where knowledge is represented as triple assertions <subject, predicate, object> statements that encode individual factual claims (e.g., <Person A, wasPromotedTo, Captain>). Making this transformation from unstructured data, particularly handwritten tabular image documents, is technically challenging. First, handwritten documents exhibit significant variability in writing style, spelling conventions, and layout organisation, making transcription errors prone \cite{Graves2000HTR}. In addition, physical degradation, such as faded ink and damaged paper, affects recognition accuracy, while irregular table layouts complicate structure reconstruction \cite{bataineh2025binarization,zhou2024enhancing,sun2025docspiral,Zhong2020PubTabNet,klut2023laypa}. Even after successful textual extraction, constructing a high-quality KG remains difficult because information extraction methods—rule-based, statistical, or LLM-driven—introduce uncertainty or inconsistency \cite{sun2025docs2kg,jarnac2024uncertainty,zhu2024llms}. Overall, the final KG often reflects a chain of multiple processing decisions whose reliability is difficult to inspect. End-to-end processes may optimise triple accuracy but obscure intermediate reasoning steps, making it impossible for users to assess why a particular assertion was generated or whether it should be trusted \cite{zhang2025trustworthy}.

This limitation is particularly problematic for humanities research from a human-AI collaboration perspective \cite{deboer2024HI}. Here, the user must be able to critically assess automatically generated triples to act on the algorithm's strengths and weaknesses; which is especially important in domains who value interpretation and accountability more than machine optimisation. The lack of transparency in AI systems has been identified as one of the main barriers to practical adoption \cite{fernandez2023fides}, in general. In historical research, data interpretations are inherently contextual, so the extracted information must remain linked to its original evidence to allow verification and multiple perspectives \cite{merono-penuela2014semantic,merono-penuela2020ontologies,ockeloen2013biographynet}. Therefore, in image-to-KG workflows, high accuracy in triple assertion alone is not sufficient: the system must support inspection, evaluation, and opportunity for correction. We operationalise those as three concrete system requirements: (1) traceability of each extracted assertion across intermediate representations, (2) evaluation of error introduced through each algorithm, and (3) preservation of data provenance to the original document.

This raises the central research question of this paper: \textit{how can knowledge graphs be constructed from handwritten tabular document images so that extracted assertions and their error propagation are transparent and traceable?} We address this problem by considering image-to-KG construction not as an end-to-end prediction task but as a provenance-aware multistep process. We introduce a modular pipeline in which table reconstruction, information extraction, and KG construction are explicitly separated, allowing each stage to produce intermediate output linked to the original image. By preserving image- and table-cell-level data provenance, the system enables users to inspect and verify machine-generated assertions and supports human-AI collaboration rather than black-box automation.

The contributions of this paper through the proposed pipeline are three-fold. First, we present a modular pipeline that converts handwritten tabular documents into knowledge graphs while exposing intermediate representations. Second, we incorporate a multi-level evaluation framework that measures performance across cell detection, table reconstruction, and semantic extraction, enabling analysis of error propagation. Third, we integrate data provenance, linking extracted entities and literals to their visual and textual evidence. 
We illustrate our proposed solution by an example implementation of this pipeline through three different variants of table reconstruction approaches and demonstrate that modular evaluation uncovers behavioural and reliability differences that would otherwise remain hidden in end-to-end systems. 
Through our case-study demonstrations on real-world archival material, we show that provenance-tracing is also possible along with automated triple assertion.  
We also demonstrate the necessity of provenance-aware pipeline through a military career event reconstruction case study, where complex, multi-step events (e.g., transfers and promotions) are error-prone, making traceability essential for inspection, validation, and correction.

\section{Background and Related Work}

Automated table reconstruction, the process of converting image-based tables into machine-understandable formats, remains a technically challenging task due to the structural variability of tables and the quality of source documents \cite{Zhong2020PubTabNet,zhou2024enhancing}. Tables embedded in PDFs and scanned images exhibit diverse layouts, implicit hierarchies, merged cells, and irregular alignments that complicate structural parsing \cite{Zhong2020PubTabNet}. While humans can intuitively infer header semantics and relational structure, automated systems must approximate these interpretations through layout analysis and pattern recognition \cite{xing2023lore}. 
Recent advances in deep learning and transformer-based architectures have substantially improved table structure recognition (TSR), and Vision-Language Models (VLMs/VLLMs) are increasingly being applied to this task \cite{zhou2024enhancing,sun2025docs2kg,kirillov2023segment}. These models demonstrate promising zero-shot and few-shot capabilities, reducing the need for handcrafted rules. However, they often operate as opaque systems, producing final structured outputs without exposing intermediate structural decisions \cite{zhang2025trustworthy}. 
This limits consistency and makes systematic error analysis difficult—an important concern in human-centred AI settings. Small transcription or segmentation errors can significantly affect downstream structural reconstruction.

In addition, input quality remains a major bottleneck for the TSR task. Blurriness, skewness, and geometric distortion significantly degrade the accuracy of table recognition \cite{zhou2024enhancing}. Although some models show robustness to missing borders, skewed or tilted tables substantially reduce performance \cite{zhou2024enhancing}, the challenge becomes more pronounced in historical and archival collections. 
Many exist only as scanned images of handwritten or degraded printed material, which require handwritten text recognition (HTR) before semantic processing can begin \cite{merono-penuela2020ontologies,klut2023laypa}. Physical degradation—such as faded ink, stains, bleed-through, and paper deterioration—further complicates legibility and extraction.

Handwritten text recognition for historical corpora remains inherently challenging due to variation in writing style, overlapping text, and inconsistent margin \cite{AlKendi2024HTR}. Even state-of-the-art systems exhibit domain sensitivity and require extensive model adaptation. Platforms such as Transkribus \cite{Kahle2017Transkribus} have become de facto standards for archival HTR workflows \cite{koert2024Loghi}, enabling large-scale transcription; however, the transcription accuracy varies between collections and directly influences the downstream structural and semantic interpretation.
Beyond HTR challenges, historical documents frequently contain irregular layouts, domain-specific terminology, and implicit contextual references that are difficult for automated systems to interpret \cite{Weber2018Towards, Ehrmann2023NER}. In such settings, errors introduced during layout analysis or HTR can cascade into structurally plausible but semantically incorrect representations. This is particularly problematic when extracted data is later formalised into knowledge graphs, where incorrect assertions may appear authoritative.

KG construction represents extracted information as structured triple assertions. While RDF-based models provide a formal mechanism for representing facts, they inherently do not capture how these facts were produced. Data provenance: the documentation of data origin, transformation processes and intermediate states—is therefore critical to ensure trust, reproducibility, and accountability \cite{kleinsteuber2024managing,jain2025enhancing}.
W3C-standardised ontologies, i.e., PROV-O and PROV-DM enable systematic representation of data provenance, linking graph assertions to source documents, workflows, and generating agents. Provenance-aware KGs support validation, reproducibility, and iterative refinement, which are central to open science practices \cite{kleinsteuber2024managing,jain2025enhancing}. They also allow for modification, rollback operations, and fine-grained inspection of individual assertions. Despite these advances, provenance is often treated as an auxiliary metadata layer rather than as a core design principle integrated throughout the extraction pipeline.

For human-AI collaboration, this gap is critical. Hybrid intelligence systems should not only optimise performance, but also enable transparency, controllability, and collaborative verification \cite{Akata2020HI}. In multi-stage document processing pipelines, error accumulates across layout analysis, HTR, table reconstruction, and semantic extraction. Without modular evaluation and explicit provenance links from high-level semantic assertions back to visual and textual evidence, domain experts lack the means to interpret system behaviour, assess reliability, or correct errors. Therefore, despite rapid progress in document AI and KG construction, there remains a disconnect between high-performance extraction systems and transparent, human-centred knowledge infrastructures. Addressing this disconnect requires architectures that expose intermediate representations, support stage-wise evaluation, and embed data provenance directly into KG construction. Such an approach aligns with the goals of Human-Centric AI by combining automated extraction with traceability and human interpretative authority, particularly in complex historical domains.

\section{Proposed Approach}
\label{sec:ch7_approach}
The task considered in this paper is: \textit{given a handwritten tabular document image, construct a knowledge graph that conforms to a predefined schema}. The requirement imposed in this task is that the generated triples are traceable to the original information in the source documents to ensure human-AI collaboration -- which is operationalised as every asserted triple must remain linked to its visual co-ordinates of the source document and textual span of intermediate (tabular) results.
To achieve this, we design a modular, provenance-aware pipeline that transforms handwritten table images into semantically structured KG. A central design principle of our system is that provenance and intermediate results are preserved at every stage of processing, ensuring that extracted entities and literals retain traceable connections back to the image regions (co-ordinates) and extracted text spans from which they were derived. The pipeline is systematically decomposed into independent components to support manual inspection, diagnostic evaluation, and step-wise verification of intermediate output. The complete workflow consists of three main stages: (1) Table Reconstruction, where the table’s structure and text content is extracted from the image; (2) Information Extraction, where row-level semantic interpretation produces structured entities and literals for property values based on given data schema; and (3) KG Construction, where assertion triples are generated together with corresponding data provenance triples. Each stage of the reconstruction process is further incorporated with evaluation metrics.

For illustrative purposes, we select an image of an archival document that contains tabular data, and use this example as a reference to clarify our pipeline description.\footnote{Image \href{https://www.nationaalarchief.nl/onderzoeken/archief/2.10.50/invnr/45@/file/NL-HaNA_2.10.50_45_0143?viewer=true}{NL-HaNA\_2.10.50\_45\_0143} from the Dutch National Archive.}

\begin{figure}[h!]
    \centering
    \includegraphics[width=1\linewidth]{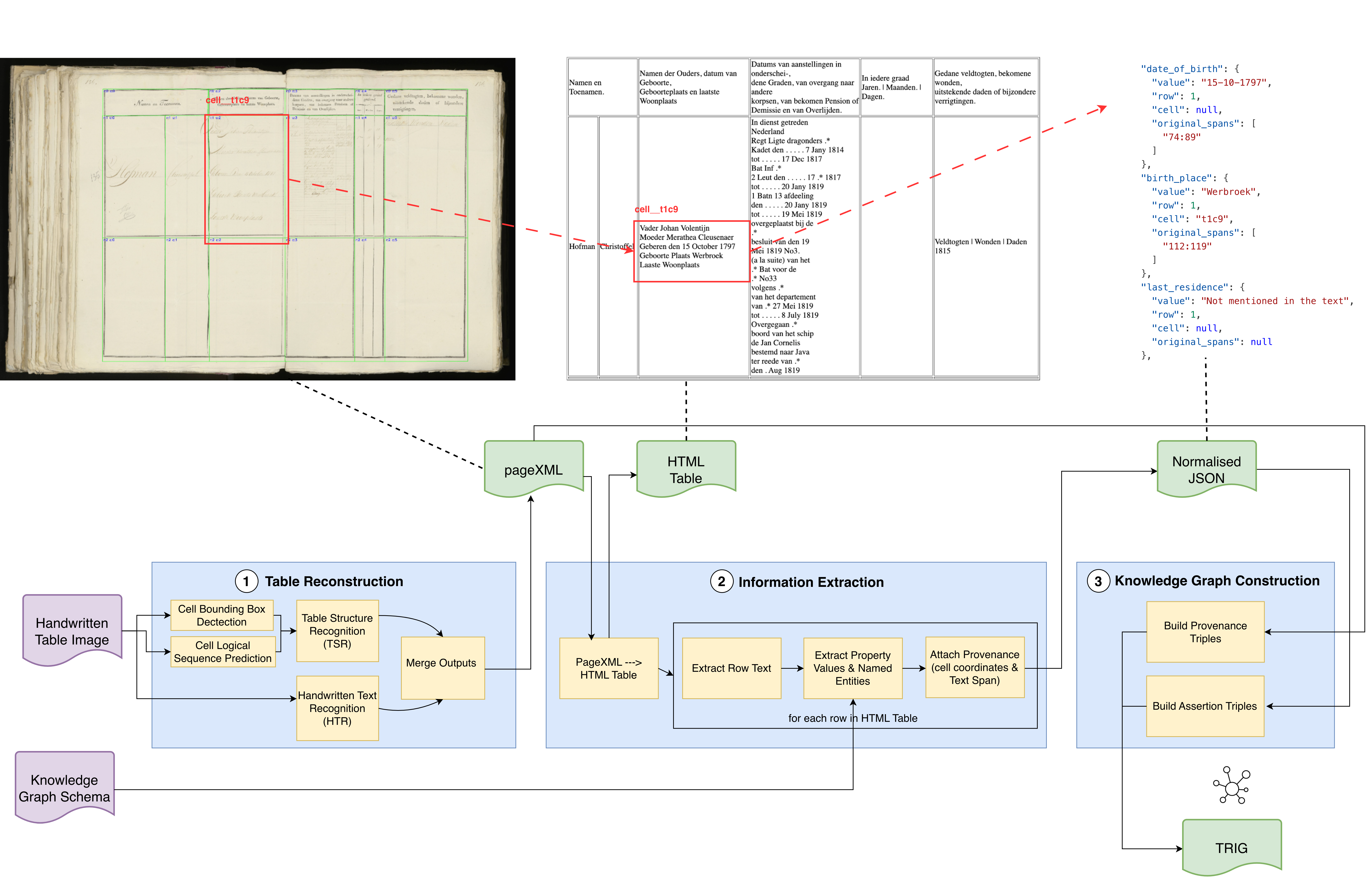}
    \caption{Illustrative architecture of our provenance-aware pipeline. Blue elements correspond to core components, yellow elements to operations within each component, purple shapes to input resources, and green shapes to generated outputs. Solid arrows indicate the direction of data flow throughout the pipeline and dotted lines visually illustrates some of the intermediate outputs.}
    \label{fig:pipeline}
\end{figure}

\subsection{Table Reconstruction} 
Given a handwritten tabular document image, the first component of the pipeline concerns table reconstruction. Our pipeline implements this component in a stepwise manner so that the resulting table representation is not merely a machine-readable abstraction of table content but a provenance-aware structure in which every cell is linked to its exact co-ordinates in the source image. This allows users to inspect where each piece of information originates and provides the opportunity for manual verification and correction whenever necessary.
The process begins with cell Bounding-box detection and table structure recognition. Here, the system identifies individual cell regions and infers their logical cell sequences, including row and column indices, as well as row spans and column spans. This corresponds to the task commonly referred to in the literature as Table Structure Recognition (TSR), which recovers the geometric grid of the table without interpreting the textual content. Therefore, the output at this stage captures only the structural layout of the table \cite{xing2023lore}.

The next step is Handwritten Text Recognition (HTR), in which the system detects text-lines and produces a transcription. The HTR component outputs a standard PageXML file, which includes both the polygonal coordinates of the recognised text-lines and their textual content. This step extracts “what is written” on the page, but does not yet assign the text to specific table cells.
To obtain a complete text-filled table, our pipeline merges the TSR-derived cell structure with the HTR-derived text-lines. This is achieved by computing the polygonal overlap between detected cell regions and text line polygons. Text-lines that overlap with a given cell are assigned to that cell, and the reconstruction process respects row spans and column spans when aggregating text. Through this mapping, the pipeline produces a structured table in which each cell contains text aligned with its spatial origin in the scanned document. The resulting PageXML document from this component serves as the intermediate machine-readable representation of the table. It embodies the complete structural and textual reconstruction and acts as the primary provenance anchor for all downstream tasks, i.e., information extraction and knowledge graph construction.

\begin{figure}
    \centering
    \includegraphics[width=0.9\linewidth]{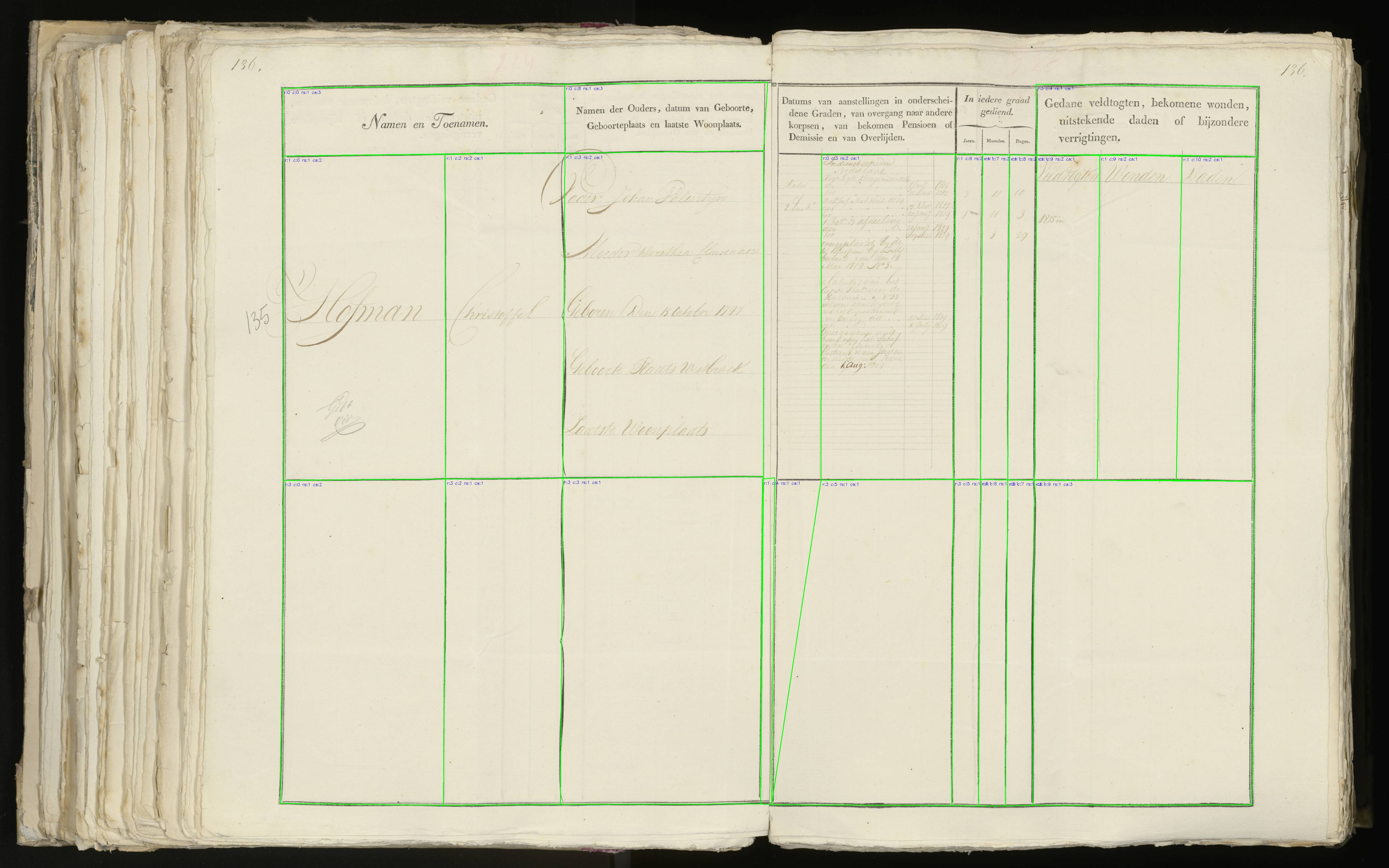}
    \caption{Automatically detected cell bounding box visualisation of image NL-HaNA\_2.10.50\_45\_0143}
    \label{fig:cell_box}
\end{figure}

A visual representation of automatically detected cell bounding boxes is shown in Figure~\ref{fig:cell_box}. The corresponding PageXML file, which records these bounding boxes and the text within them, can be found in the github repository\furl{PageXML for image NL-HaNA\_2.10.50\_45\_0143}{https://github.com/Shoilee/stamboeken_htr/blob/main/examples/NL-HaNA_2.10.50_45_0143.jpg.xml}.

\subsubsection{Evaluation Metrics} For the first operation in the Table Reconstruction component, cell detection, we employ the mean Average Precision (mAP) \cite{everingham2010pascal}, which summarises the precision-recall performance across confidence thresholds. mAP allows us to assess how accurately a cell detection method identifies the image region for the given table cells. To evaluate the precision of the overall reconstructed table, we use Tree Edit Distance (TED) \cite{pawlik2016TED} and its text content ignorant variant TED-Struct \cite{zhou2024enhancing}. TED quantifies the minimal number of edit operations required to transform the predicted table into the ground-truth one, capturing structural errors such as missing or incorrect splits and merges and also mismatches of text alignments. TED-Struct focuses only on the structural alignment of the table, ignoring the textual content. More details on how these metrics can be implemented in practice are provided in the Appendix~\ref{app:tsr_illustration} with the illustrative running example.


\subsubsection{Implementation Specification}
\label{sec:tsr_implement}
To demonstrate the possibility of implementing different approaches, we implemented three table reconstruction variants drawn from distinct methodological families. Variant-1 used Transkribus, the proprietary domain standard. Variant-2 integrates different deep learning based pre-trained models for segmented tasks. Variant-3 used a large language model (LLM) for table reconstruction.
Each variant followed the same core sequence of steps: (1) detecting cell bounding boxes, (2) generating logical cell sequence or table structure recognition, (3) performing handwritten text recognition (HTR), and (4) finally producing a machine-readable table by aligning the outputs of these individual stages. The exact implementations of each of these variants are given below: 
\\
\\
\noindent\textbf{Variant-1} To incorporate the Transkribus-based table reconstruction variant, we used the industry-standard Transkribus~\cite{Kahle2017Transkribus} platform through its user interface. Transkribus is the de facto industry standard for historical document analysis, offering state-of-the-art HTR models trained on diverse European handwriting corpora, robust layout analysis for tables/columns, and a mature ecosystem with community-contributed models. 18 images were randomly selected from our target dataset to fine-tune the existing table recognition model, which was later evaluated on two unseen images, achieving a mean Average Precision (mAP) score of 0.48. It is worth noting that both the training and validation images shared a similar table layout -- a recommended practice for optimal performance of the table recognition feature. Following table recognition, handwritten text recognition was performed using the Dutch Handwriting M1 model (CER= 0.0510), contributed by the Transkribus Community.
\\
\\
\noindent\textbf{Variant-2} We implemented an ensemble of existing pre-trained models for each individual task in the table reconstruction pipeline. The LORE Wired model from paper \cite{xing2023lore} was used for both cell bounding box detection and table structure recognition; this model was trained on tables featuring visible cell-separating lines from SciTSR \cite{chi2019SciTSR} and PubTabNet \cite{Zhong2020PubTabNet} dataset, allowing it to effectively predict spatial and logical location of table cells. For HTR, we used the Loghi tool \cite{koert2024Loghi}, using its publicly available base model with CER below 3\% on Dutch material from the 17th century. Loghi first identifies the baseline of each text line to group sentences correctly, then Loghi converts the text image into digital text. To generate the final PageXML output, we computed the polygonal overlap between each text line detected in the HTR output and the corresponding cell bounding boxes from the table structure recognition (TSR) stage. A text line was assigned to a cell if their overlap area ratio was $\ge$ 0.2.
\\
\\
\noindent\textbf{Variant-3} We implemented this variant to evaluate whether the proposed pipeline can be realised using vLLM. For that, we used the multi-run conversation function, where we asked the agent to perform individual steps in each conversation. First, we asked the agent to recognise cell-bounding boxes and construct logical cell sequences. Then, in the next conversation turn, we asked the agent to do handwritten text recognition, and finally in the last conversation turn, we asked to combine the output from construct the pageXML file in the given format. We stored all the intermediary outputs and also the final pageXML that records table structure with each cells textual content along with its co-ordinate. 
The prompt used for each of these tasks can be found here\furl{Table Reconstruction Prompts}{https://github.com/Shoilee/Image2Table-reconstruction-with-llm/blob/main/prompt.py}. All prompts were executed using the vision model \verb|meta-llama/llama-4-scout-17b-16e-instruct| accessed via the Groq API\furl{Groq Api}{https://console.groq.com/docs/api-reference}.

\subsection{Information Extraction}
\label{sec:IE}

The information extraction component focuses on identifying and extracting property values for each entity of interest, given a predefined KG schema. Here, we make the assumption that each table row corresponds to a single entity of interest; therefore, extraction is performed row by row, and the row-wise results are concatenated at the end to obtain the full image-level structured information.
In this component, the pipeline begins by converting the PageXML table output from the previous stage into an HTML table. This conversion supports both human readability and efficient row-level iteration. For each row, the pipeline concatenates the text content of its constituent cells (columns), providing richer contextual information than isolated cell-level fragments. The choice is justified by initial experiments showing that row-level extraction consistently outperforms cell-level extraction, mainly because it is more robust to localised text errors within individual cells. However, this approach temporarily withdraws cell-level provenance, which is later restored by mapping text-span indices back to the corresponding cells. As a result, extracted named entities or literals can still be associated with the cell provenance, provided that they correspond to the source text.  The final output of this component is stored as a normalised JSON object (one per image), which is a list of entities from each table row with their schema-guided property value along with their associated provenance metadata (row index, cell ID and text spans).

\begin{figure}
    \centering
    \includegraphics[width=1\linewidth]{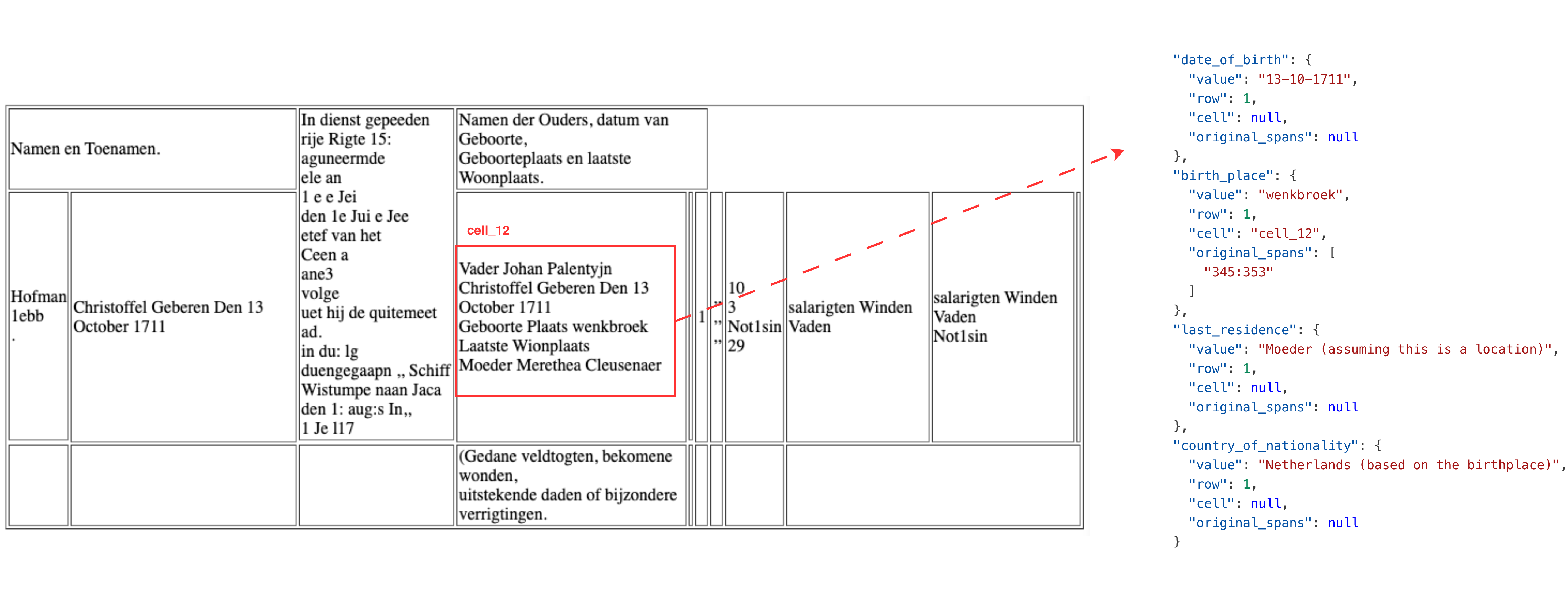}
    \caption{On the right, the image displays a snippet of the structured information extracted from the HTML table (in the left), together with its corresponding cell- and text-level provenance, where available. Note that, some cell provenance is lost when the extracted information cannot be directly (string) matched to the original cell text as explained in Section~\ref{sec:IE}.}
    \label{fig:info_ext}
\end{figure}

Figure~\ref{fig:info_ext} shows the  extracted information given the pageXML table of the running example. The corresponding JSON file for this image, which records both structured information and cell and/or text provenance (when available), can be found here\furl{PageXML for image NL-HaNA\_2.10.50\_45\_0143}{https://github.com/Shoilee/stamboeken_htr/blob/main/examples/NL-HaNA_2.10.50_45_0143.jpg.json}.

\subsubsection{Evaluation Metrics} The pipeline incorporates evaluation metrics such as precision, recall, and F1-score for information extraction from each image.  Again, the practical implementation of these metrics is illustrated in Appendix~\ref{app:ie_illustration} using a concrete example. 

\subsubsection{Implementation Specification}
\label{sec:ie_implement}
As described in Section~\ref{sec:IE}, we first reconstruct an HTML table from PageXML output, then, for each logical row, concatenate cell texts to provide the IE system with more complete information. In contrast to the first component , only a single variant is implemented for this step. However, this component can be replaced by any text-to-KG extraction method. For Information Extraction, here we employ OntoGPT \cite{caufield2024structured} as an off-the-shelf, schema-guided information extraction module, running through the Ollama framework with the Llama3 model (ollama/llama3). OntoGPT produces YAML output that contains predicted entities, property values, and text-span annotations, given a desired schema. For each predicted literal or named-entity value, we record the character‐offset span and map it back to the originating cell identifier from the HTML table cell. However, when the extracted property or attribute value does not exactly match the corresponding string in the source text, the cell-level provenance for that value is lost. This is a limitation of our current implementation. Ultimately, the overall process allows us to attach every property value to its table cell ID and, consequently, to the corresponding bounding-box coordinates in the image, whenever possible . In the end, the system produces a JSON file containing one or more predicted entity records per image document.

\subsection{Knowledge Graph Construction} 
The final stage of the pipeline converts the normalised JSON output into a KG.  At this point, each extracted entity from the information-extraction component is represented as a JSON object containing (a) its property and attribute values and (b) fine-grained provenance annotations linking every value to its original text span, cell ID, and row index. Using the cell ID, the system retrieves the corresponding image coordinates from the earlier PageXML output. 
This KG Construction component has two goals: (1) to generate RDF assertions aligned with the given schema (assertion graph), and (2) to generate RDF triples based on the provenance information captured in the preceding steps ensuring traceability to the source image document and intermediate table (provenance graph).
\begin{figure}[h!]
    \centering
    \includegraphics[width=0.6\linewidth]{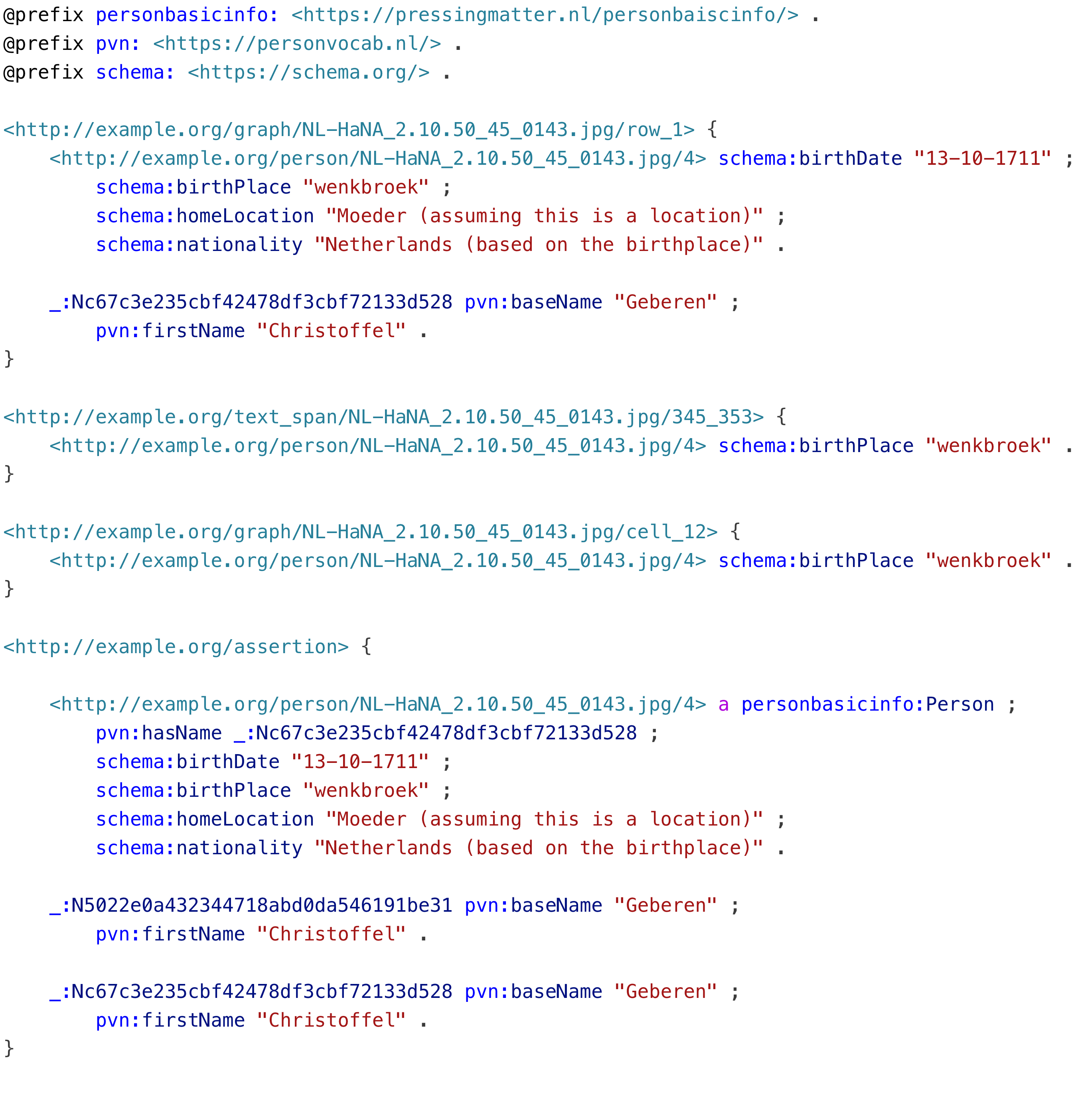}
    \caption{Snippet of assertion graph for image NL-HaNA\_2.10.50\_45\_0143}
    \label{fig:assert_graph}
\end{figure}
For each entity, the system first builds the assertion graph by minting URIs, assigning entity types, and mapping property assertions according to the schema. To maintain traceability, every triple is also added to three provenance-specific named graphs: row-level (e.g., :prov\_row\_1), cell-level (e.g., :prov\_cell\_12), and text-span (e.g., :prov\_span\_241\_250). Each links the assertion to its corresponding row, cell, and text segment. In parallel, a provenance graph records the details of these named graphs. For every named\_graph in the assertion graph, we create a provenance node that details the offset of the originating text range, the associated table row index, or the cell’s bounding-box coordinates in the PageXML representation. This follows standard provenance modelling practices (e.g., PROV-O\furl{PROV-O}{https://www.w3.org/TR/prov-o/}), where each assertion is linked to its evidence through \verb|prov:wasDerivedFrom| or equivalent relations. This design ensures that the KG encodes both extracted knowledge and a transparent audit trail for inspection, validation, and correction. 

\begin{figure}[h!]
    \centering
    \includegraphics[width=0.8\linewidth]{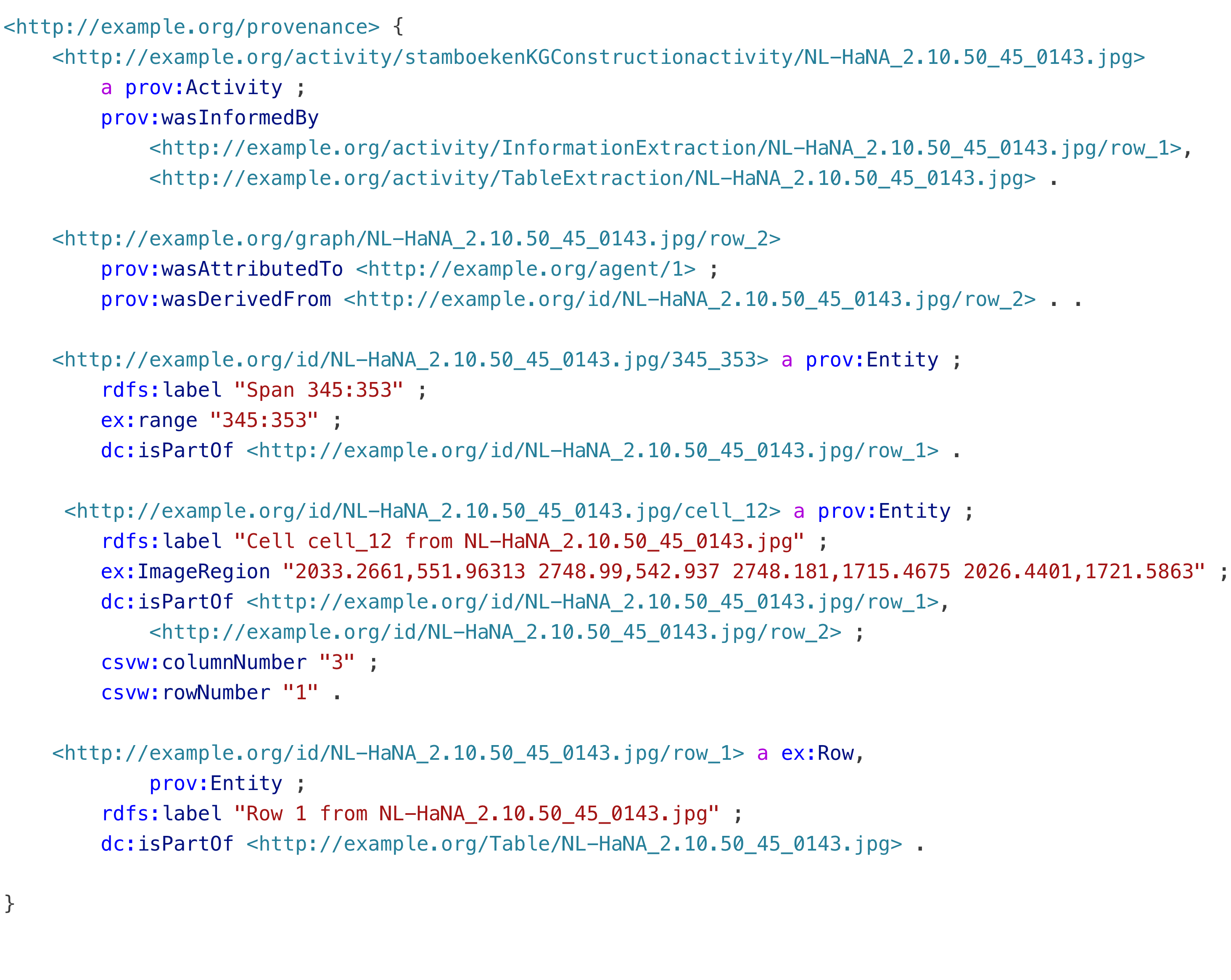}
    \caption{Snippet of provenance graph for image NL-HaNA\_2.10.50\_45\_0143}
    \label{fig:prov_graph}
\end{figure}

Figure~\ref{fig:assert_graph} shows the assertion graph in which triples are stored within one or multiple named graphs depending on their data provenance. These named graph URI's are further used in provenance graph to describe more provenance details about such assertion, shown in Figure~\ref{fig:prov_graph}.
\\
\subsubsection{Evaluation Metrics} As the assertion graph is constructed via a direct mapping from the structured information produced by the previous component, we assume that precision, recall, and F1-score directly reflect the semantic correctness of the resulting assertion graph. The consistency of the provenance graph is evaluated using SHACL shapes. We define a set of SHACL shape constraints that specify the required structure and properties for provenance triples as described above. The data provenance SHACL shape specification for the current pipeline is available here~\furl{Data provenance SHACL shapes}{https://github.com/Shoilee/stamboeken_htr/blob/main/data/schema/data_provenance.ttl}. Given the constructed provenance graph and the corresponding SHACL shapes, a SHACL validator is used to verify whether all nodes conform to the specified shapes. The validator returns a Boolean result, where the output is `True' only if every node satisfies its associated shape, and `False' if at least one node violates any constraint.

To further assess the extent of cell-level provenance preserved by the pipeline, the system computes descriptive statistics over the resulting provenance annotations, i.e., number or ratio of properties per entity with cell provenance. In addition, to evaluate the semantic correctness of triples enriched with cell-level provenance, the pipeline independently reports precision, recall, and F1-score as quantitative performance metrics.

\subsubsection{Implementation Specification}
The final KG construction comprises two complementary steps. First, the normalised JSON output is transformed into an RDF \textbf{assertion graph} using a custom Python function \verb|build_assertion_graph|, available in the KG construction script~\furl{KG conversion script}{https://github.com/Shoilee/stamboeken_htr/blob/main/src/constructPersonBasicInfoKG.py}. This function recursively traverses flat and nested JSON structures, systematically converting each schema-guided property-value pair into conforming RDF triples while preserving full provenance metadata (row index, cell ID, text spans) within named-graph contexts or reified statements.
Second, a parallel function generates the \textbf{provenance graph} following the data provenance schema defined in Section 3. For each assertion triple, it creates provenance triples linking the semantic statement to its originating cell coordinates, text spans, and table structure, following standard PROV-O conventions. The resulting provenance graph is then validated against the provenance SHACL shapes to ensure structural consistency and schema conformance.

The final output is an integrated KG that combines the assertion graph and the provenance graph. This representation supports downstream querying, integration with broader domain ontologies, and manual validation workflows. Importantly, explicit capture of cell-level provenance enables granular inspection of extraction errors and facilitates high-confidence data curation, reinforcing the pipeline’s design goal of transparency and trustworthiness.
The complete pipeline implementation, including all source code, schemas, configuration files, and example data, is publicly available in our project repository~\furl{Pipeline repository}{https://github.com/Shoilee/stamboeken_htr/}.


\section{Pipeline Demonstration}
\label{sec:ch7_eval}
To validate the pipeline and its modular design, we conduct three complementary demonstrations that cover both granularity and scale. First, we analyse a small set of five documents using gold-standard annotations and an in-depth demonstration focusing on simpler biographical data extraction, which allows us to inspect the behaviour of individual pipeline components in detail. Second, we demonstrate the pipeline on a larger collection of 3,000 triples as gold-standard end triples, evaluating only the end-to-end triple assertion performance with provenance preservation and again restricting the scope to biographical information, in order to assess robustness and scalability under realistic large-scale processing conditions. Third, we examine a single document in a more complex event-centric setting, military career events extraction (e.g., transfers and promotions), to demonstrate how the modular pipeline supports fine-grained inspection, correction, and provenance-aware event reconstruction in order to allow human-in-the-loop.

To demonstrate the pipeline, we conducted experiments using real archival material. Our source documents are the service and pension records of the army in the Dutch East Indies (KNIL) and of the troops in the West Indies, 1815–1951. These registers contain handwritten tabular data with biographical and career information about military personnel. This archive series is organised into `folios' (books),  each consisting of around 300–600 pages, and in total there are 761 such books; which are publicly available\furl{Source documents}{https://www.nationaalarchief.nl/onderzoeken/archief/2.10.50}. For all of our experiments, we focused on images from one single book, folio-45.

\subsection{Demonstration with Modular Pipeline}
\label{exp:modular_eval}
\noindent\textbf{Set-up. }
For in-depth demonstration of our modular pipeline, we first implemented three distinct variants of the Table Reconstruction component with one variant of our Information Extraction component, as described in Section~\ref{sec:tsr_implement} and Section~\ref{sec:ie_implement}. In addition, we implemented two baseline approaches:  
(1) \emph{Baseline}: a straightforward layout-to-text conversion based on large language models (LLM), in which the cell bounding box detection steps within the Table Reconstruction component were omitted and prompted the LLM to generate an HTML table given an image. This baseline enabled us to assess whether explicit cell bounding box detection provides benefits for table reconstruction and information extraction beyond its role in provenance tracking.  
(2) \emph{GT HTML}: a configuration in which we skipped the Table Reconstruction component entirely and instead supplied manually curated ground-truth (GT) HTML as system input. This setting allows us to evaluate the performance of only the Information Extraction (IE) component and thereby analyse the performance drop in IE performance that occurs when tables are automatically generated.

For this demonstration, we used five randomly selected images from folio-45, each paired with manually curated ground truth for both cell bounding boxes and the corresponding expected (true) HTML table (GT HTML). Table reconstruction performance was measured using mAP (cell detection precision), TEDS-Struct (table structure similarity), and TEDS (table structure+content similarity), allowing us to perform a comparative analysis between three different variants of Table Reconstruction and discuss the specific strengths and weaknesses of the implemented variants. Downstream information extraction was reported using precision, recall, and F1-score. The score reported in Table~\ref{tab:result_exp1} is averaged over five sample images. The exact implementation of these evaluation metrics can be found in Appendix~\ref{app:tsr_illustration} and Appendix~\ref{app:ie_illustration}.
\\
\\
\noindent\textbf{Result Analysis. }
The results in Table~\ref{tab:result_exp1} highlight three key findings. First, the three different table reconstruction methods exhibit substantial performance differences in modular performance. For Variant-1 and Variant-3 all table reconstruction metrics are higher than for Variant-2. However, it is the opposite when it comes to IE performance. Qualitative observation on this analysis shows that this discrepancy is caused due to the HTR error. Although both both TED-struct score and the TED score inform about the performance in table reconstruction, the difference in their score was due to the HTR error. This discrepancy contributed quite significantly to the IE performance for Variant-1, although it had a nearly perfect TED-struct and maP score. We conclude here that good table structure does not necessarily guarantee strong IE performance, particularly when affected by handwriting recognition quality, demonstrating the value of a modular pipeline that allows comprehensive comparison of intermediate components. 

\begin{table}[t!]
\scriptsize
    \centering
    \caption{
    Comparison of three implemented handwritten table reconstruction variants and their downstream impact on information extraction (IE). 
    Structural metrics (mAP, TEDS, TEDS-Struct) evaluate the quality of table reconstruction, while IE precision/recall/F1 capture KG extraction performance. 
    The \textit{Baseline} column shows IE results using an LLM-generated table without modular cell detection or HTR, and thus without image-level provenance. 
    The \textit{GT HTML} column reports IE performance on the hand-annotated ground-truth table, serving as an upper-bound reference for the IE component under perfect table reconstruction.
    }
    \label{tab:result_exp1}
    \begin{tabular}{{|c|c|c|c||c|c|}} \hline
                           & Variant-1  & Variant-2  & Variant-3   &  Baseline & GT HTML \\ \hline
         maP score         & 0.9444      & 0.5454      & 0.0000       &     --    & 1.0     \\ \hline
         TED-struct score  & 0.9632      & 0.7650      & 0.8782       &   0.9048  & 1.0     \\ \hline
         TED score         & 0.8429      & 0.6777      & 0.7442       &   0.7870  & 1.0     \\ \hline
         IE Precision      & 0.0222      & 0.3319      & 0.2182       &   0.0778  & 0.5756  \\ \hline
         IE Recall         & 0.0308      & 0.3095      & 0.2001       &   0.0816  & 0.6605  \\ \hline
         IE F1-score       & 0.0258      & 0.3200      & 0.2066       &   0.0792  & 0.6449  \\ \hline
    \end{tabular}
\end{table}

The second important observation comes from the behaviour of Variant-3. Its mAP score is exactly zero, despite achieving reasonable TED and TED-Struct scores. Because our LLM-based implementation separates the prediction of the bounding-box from the generation of the table structure, and we evaluated each part independently. The mAP score reveals that none of the predicted cell coordinates overlap with the ground truth—demonstrating that the LLM fabricated bounding boxes coordinates rather than grounding them in the actual document image. This fabricated spatial provenance became detectable due to the modular evaluation. 


A third key finding is the effect of error propagation across the pipeline. Table~\ref{tab:result_exp1} showing that the downstream IE performance is highly sensitive to both the table reconstruction and the HTR quality. A perfect structure alone does not ensure better extraction; e.g., Variant-1 achieves near-perfect table reconstruction performance but poor IE due to transcription errors (gaps in between TED and TED-struct score), while Variant-2 performs best overall by producing cleaner text despite weaker structural metrics. Moreover, when the IE component applied to a manually constructed ground-truth table (GT HTML) shows some performance loss, underscoring that errors can accumulate at every stage. This result highlights that automated information extraction remains constrained by the complexity of the task, emphasising the need for human–machine collaboration to achieve reliable results. 

In addition, the end-to-end table reconstruction approach using LLM (Baseline) exhibits lower performance compared to Variant-3, which is also an LLM-based method. In Variant-3, however, the table is reconstructed through a multi-step procedure: in the first step, the LLM is prompted to reconstruct cell bounding boxes and infer the logical row sequence; in the second stage, it is prompted to perform handwritten text recognition; and in the final stage, it is asked to concatenate the outputs from the previous stages to generate the HTML table. Comparison with this baseline indicates that reconstructing tables via such a modular, multi-step process confers clear advantages over a purely end-to-end strategy.

\subsection{Demonstration on End-to-End KG Construction} 

\noindent\textbf{Set-up.} To scale-up to larger numbers of documents, we used the ground truth triples from the Colonial Collections Hub, Stamboeken KG, a publicly available knowledge graph\furl{Stamboeken KG}{https://data.colonialcollections.nl/Bronbeek/Stamboeken}\footnote{Note that, (at the time of writing), it is temporarily restricted due to licencing issues.}. This knowledge graph was developed as part of a separate initiative to allow the analysis and reuse of military personnel data from the colonial-era. The triples were derived from an excel dataset created through manual data entry, capturing basic biographical attributes such as name, birth and death details, places of birth, and last known military rank. The curated dataset was then converted into a structured knowledge graph to support semantic linking and interoperability within the Colonial Collections Hub.




For triple assertion performance comparison by the pipeline, we randomly sampled 91 table images from  folio-45 which consist of 98 unique individuals person and 3000 triples. The pipeline generated triples were compared with these samples as ground-truth reference data to assess downstream KG construction performance. To support image-level evaluation, ground-truth triples were grouped by source image and stored as normalised JSON files—one per image—containing all entities and attributes for that page. 

Subsequently, we computed a Hungarian alignment \cite{kuhn2005Hungarian} for each image between the predicted and ground-truth entities to determine the optimal alignment. For each matched entity pair, property values are compared using string similarity; values with similarity $\geq$ 0.6 are considered correct. Literal properties and named-entity properties (including their own nested attributes) are equally incorporated into the scoring. 
We report precision, recall, and F1-score for automated triple construction to evaluate how different table-reconstruction approaches affect downstream information extraction (IE) performance.

We further report the descriptive statistics about the identified person instances using different variants of table reconstruction implementations. In particular, we quantify the number of properties (triples) that were asserted, the property–instance ratio, the number of properties supported by cell-level provenance, and the ratio of cell-level provenance–backed properties to the total number of properties.

In addition to triple-level accuracy and descriptive statistics about the assertion graph, we also measure how successfully each method preserves fine-grained provenance. Specifically, we measure how many extracted property values can be mapped back to a cell-level origin in the reconstructed table and whether having cell-level provenance contributes to better IE performance. So, we also computed the precision, recall and F1-score of IE restricted only to properties with valid cell-level provenance.
\\
\\
\begin{table}[b!]
\scriptsize
    \centering
    \caption{Comparison of the three table-reconstruction approaches in terms of end-to-end knowledge-graph assertion performance over 91 images. The first three rows report performance in terms of information precision, recall, and F1-score. The subsequent five rows present descriptive statistics of the constructed assertion graph, including its ability to preserve fine-grained, cell-level provenance. The final three rows repeat the information-oriented metrics, but this time restricted to properties for which cell-level provenance is available.}
    \label{tab:result_exp2}
\begin{tabular}{|c|l|c|c|c||c|}
    \hline
    & & Variant-1 & Variant-2 & Variant-3 & Expected (GT) \\ \cline{1-6}
    \multirow{3}*{\begin{tabular}[c]{@{}c@{}}Assertion triples\end{tabular}} & Precision & 0.12 & 0.27 & 0.12 &  \\ \cline{2-6}
    & Recall & 0.09 & 0.21 & 0.08 &  \\ \cline{2-6}
    & F1-score & 0.11 & 0.23 & 0.10 &  \\ \hline \hline
    \multirow{5}*{\begin{tabular}[c]{@{}c@{}}Assertion graph\\statistics\end{tabular}} & Total person instances & 140 & 288 & 129 & 97 \\ \cline{2-6}
    & Total properties (with non-empty values) & 825 & 1731 & 716 & 690 \\ \cline{2-6}
    & No. of properties/instance & 5.89 & 6.01 & 5.55 &  7.11 \\ \cline{2-6}
    & Total properties with cell provenance & 161 & 420 & 129 & 690 \\ \cline{2-6}
    & No. of cell provenance/properties & 0.18 & 0.25 & 0.13 &  1.0 \\ \hline \hline
    \multirow{3}*{\begin{tabular}[c]{@{}c@{}}Assertion triples \\with cell \\provenance\end{tabular}} & Precision & 0.14 & 0.39
 & 0.07 &  \\ \cline{2-6}
    & Recall  & 0.03 & 0.09 & 0.01 &  \\ \cline{2-6}
    & F1-score & 0.05 & 0.14 & 0.02 &  \\ \hline
\end{tabular}
\end{table}

\noindent\textbf{Result Analysis. }
As expected, the performance of the information extraction deteriorates further as the scale increases (Table~\ref{tab:result_exp2}), consistently across all variants of table reconstruction. However, the performance at this larger scale remains in line with the earlier results obtained using five images. However, when considering only precision, recall, and F1-score, it is difficult to identify the underlying causes of the performance differences between the various table reconstruction variants. To explain these variations, it is necessary to refer back to the more fine-grained analysis in Table~\ref{tab:result_exp1}.

Examining the assertion graph statistics, we see that the numbers of detected instances and properties are far higher than expected. The count of identified instances is directly correlated with the TED-struct scores of the tsr variants, because the pipeline assumes exactly one instance per row; therefore, a larger number of (incorrectly) detected rows leads to more person instances. Note that though for this count, we excluded any empty instances (instances without property values). The higher instance count in turn results in a higher total number of detected properties. Again, for the total property count, we removed any empty attribute values (for example, cases where the value contains phrases such as `not mentioned'\footnote{This comes from OntoGPT, which is an LLM-based information extraction method.}).

Table~\ref{tab:result_exp2} also shows that the three reconstruction approaches differ not only in the number of property values they extracted overall, but more importantly in how effectively they preserve cell-level provenance, meaning how much their result is grounded on the reconstructed table. This is essential for transparency, human correction, and provenance-aware KG construction. The results reveal that a substantial portion of the properties extracted across all the methods cannot be assigned to a specific cell. This is also a limitation of our implementation choice as IE was performed on row-level text, some extracted values are inferred from context or generated by the OntoGPT model which sometimes are not a exact literal match to any cell text. As a result, only a subset of property values have a valid cell-level provenance, and therefore only these can be used to assess the ``provenance-preserving” accuracy.

Even when a property value can be traced to a table cell, it does not guarantee semantic correctness of the property value, or it does not necessarily increase precision. The low precision scores in triples with cell-provenance in Table~\ref{tab:result_exp2} show this. Here, errors arise for three main reasons: 
(1) the IE component may misidentify the property–value pair;
(2) the HTR model may introduce transcription errors; or
(3) the TSR output may mis-segment or mis-align cells, assigning a cell to the wrong row.
This reinforces why cell-provenance alone is not enough—the correctness of upstream components directly shapes downstream extraction reliability.


\subsection{Demonstration on Complex 
\label{sec:ch7_demo}
Information Extraction}

While our previous demonstrations focused on simpler information extraction tasks to demonstrate the pipeline, we now showcase its practical value through a concrete case study: military career event extraction from handwritten archival tables. Event detection in historical documents is inherently complex, requiring disentanglement of multi-step narratives embedded in dense archival documents~\cite{sprugnoli2019novel}. Previous work shows that historical research gains maximum insight when data is viewed through an event lens—as exemplified by CIDOC CRM's emphasis on dynamic occurrences over static facts~\cite{doerr2006documenting}.

\begin{figure}[ht!]
    \centering
    \includegraphics[width=1\linewidth]{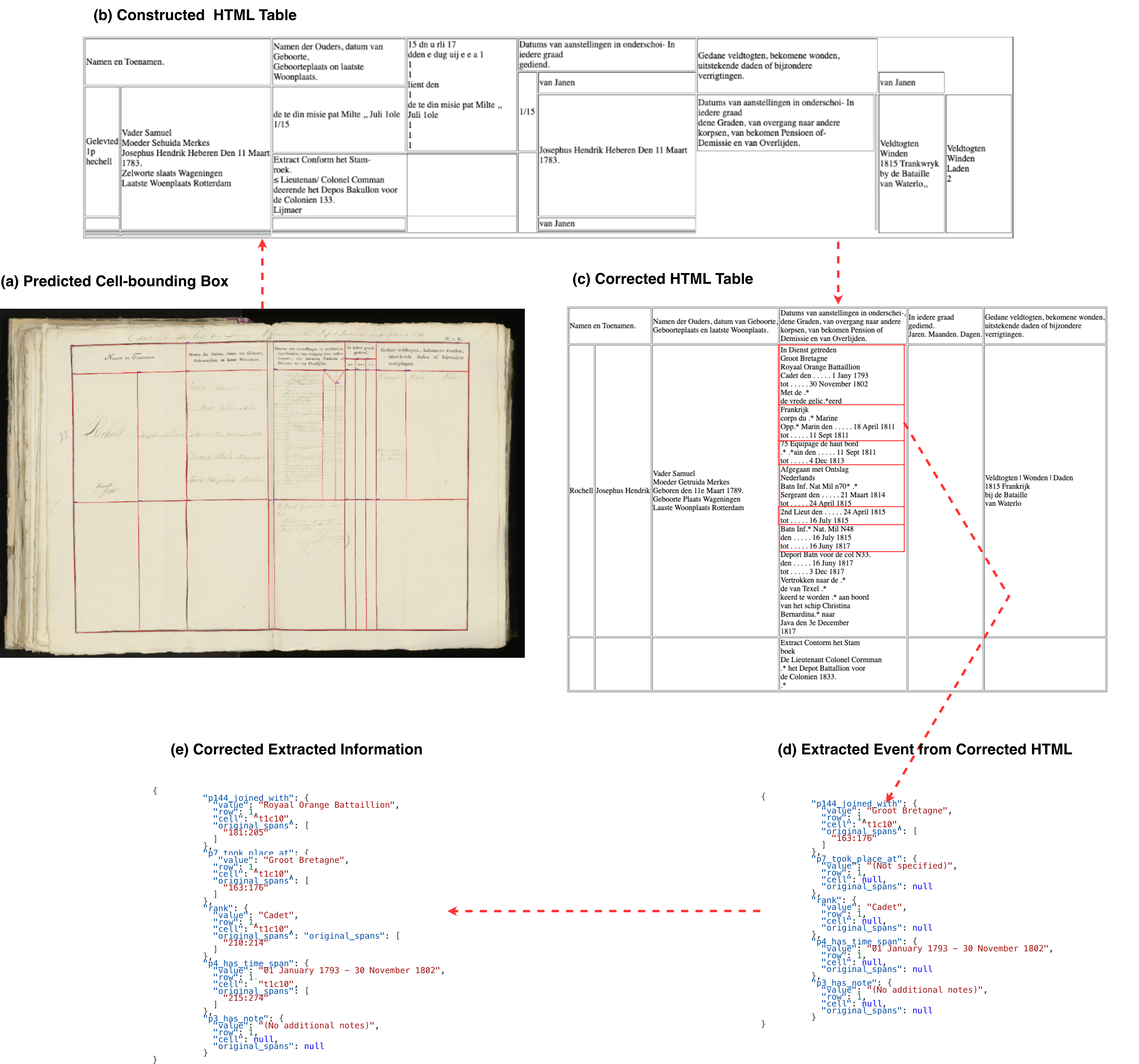}
    \caption{Event extraction from a degraded handwritten military career register using proposed pipeline and human correction.
        (a) Predicted cell bounding boxes (maP=0.1),  
        (b) Reconstructed HTML table (with TED=0.549, TED-struct=0.622), 
        (c) Human-corrected HTML, 
        (d) Pipeline-extracted events 
        (e) Human-corrected events,
    }
    \label{fig:event}
\end{figure}

Military biography reconstruction offers an important use-case: tracing personnel movements across troops reveals potential involvement in colonial campaigns and cultural heritage collections. However, manually processing thousands of digitised career registers requires excessive effort. At the same time, the automated generation of such information remains a non-trivial challenge, making it an ideal case for humans-AI collaboration. We therefore conduct a deep-dive analysis of a single representative image (Fig.~\ref{fig:event}) from our current low-quality archival sources characterised by faded ink, overlapping layout, and densely packed text.

This case study demonstrates how our modular, provenance-aware pipeline can enable human-AI collaboration for event extraction. We repurpose the WarSampo ontology~\furl{Adapted military posting schema}{https://github.com/Shoilee/stamboeken_htr/blob/main/data/schema/militarypostingevents.yaml}~\cite{koho2021warsampo} for military posting events . Through fine-grained inspection—at cell detection, HTML reconstruction, and semantic extraction stages—historians can correct errors (segmentation, HTR, event boundaries) that propagate automatically to the final KG. Provenance links from image spans to event triples ensure scholarly traceability, transforming challenging sources into reliable, event-structured knowledge graphs via hybrid intelligence.

Figures~\ref{fig:event} illustrate the workflow. Although the predicted cell bounding boxes (\textit{Fig~\ref{fig:event}(a)}) are reasonably accurate, the reconstructed HTML table (\textit{Fig~\ref{fig:event}(b)}) nonetheless does not retain any event information  (with TED and TED-struct scores of 0.549 and 0.622, respectively). This breakdown arises from two main causes: (1) layout complexity, in which overlapping strokes, text overflow, and implicit nested structures complicate the printed cell borders; and (2) low HTR performance, as faded handwriting remains challenging even for domain-adapted models, rendering substantial portions of the text unrecoverable. Manually correcting the HTML table (\textit{Fig~\ref{fig:event}(c)}) is relatively straightforward for structural reconstruction, although certain textual content cannot be recovered even by a human (author) annotation (denoted by asterisks: .*), illustrating the difficulty of the text recognition task. After this correction, the IE module identifies 7 distinct events (available here\furl{Aumatically detected events from GT HTML}{https://github.com/Shoilee/stamboeken_htr/blob/main/examples/events.json}), although with some missing or incorrect attributes (\textit{Figure~\ref{fig:event}(d)}). Through additional manual refinement of the semantic annotations (\textit{Figure~\ref{fig:event}(e)}), these events can be polished and integrated into a provenance-aware KG, optimising the distribution of efforts: automated methods tackle scalable structure detection, while human experts resolve interpretive uncertainties. In such low-quality, high-complexity scenarios, end-to-end automation cannot produce reliable KGs without human-in-the-loop intervention. Given the sensitivity of military career data, fine-grained provenance—from image spans to event attributes—is essential to ensure authenticity, enable scholarly scrutiny, and prevent misrepresentation.

\section{Discussion}
This evaluation has shown that a modular, provenance-aware pipeline can make information extraction from complex handwritten archival tables both feasible and analysable in a systematic way. By separating cell detection, table structure recognition, handwriting recognition, and schema-guided extraction, we were able to quantify how each component—and each modelling choice—affects downstream knowledge graph construction, rather than treating the system as an opaque end-to-end black box. Our analysis of cell-level provenance further demonstrated that fine-grained links back to individual table cells are crucial for transparency, error diagnosis, and human correction, but they do not in themselves guarantee semantic correctness. This paper emphasises that errors can accumulate at every stage—from TSR and HTR to IE—so meaningful evaluation and correction require component-level metrics and explicit intermediate representation. The modular design we propose supports exactly this kind of targeted diagnostic evaluation through intermediate representation. Our study also highlights both the promise and the limits of automation in historical settings. Even when provided human annotated HTML tables, the IE component does not achieve perfect performance due to the semantic complexity, variability, and noise inherent in historical documents. We therefore argue that the most robust path forward is not fully automatic extraction, but human–machine collaboration: machines provide scalable, provenance-rich candidate structures and triples, while human experts validate, correct, and extend them. The military career event extraction case study (Fig.~\ref{fig:event}) exemplifies this: despite moderate success of TSR (TED=0.549), severe degradation forced human intervention. In low-quality, high-complexity scenarios, end-to-end automation often fails, so fine-grained provenance from image spans to event attributes is essential for scholarly validation of sensitive historical data.

Note that, this evaluation did not seek to identify the single best table reconstruction method, but rather to test whether IE metrics alone provide sufficient insight for algorithmic selection. Our results reveal that aggregate scores mask component-specific failure and consequently impacts on KG quality. This work prioritised feasibility over optimising any single component, validating that the proposed pipeline exposes diagnostic signals for targeted intervention. Future research can leverage this infrastructure to systematically compare TSR methods on handwritten archival tables, attributing performance not just to end-to-end scores but to their true contribution across the full image-to-KG workflow. 

Our future work will prioritise the development of human-in-the-loop interfaces that communicate intermediate representations—such as detected cells, reconstructed tables, extracted triples, and provenance trails to the user, enabling targeted user interventions. A particularly relevant direction for subsequent work is to empirically investigate whether such modular communication mechanisms effectively foster user trust. 
In addition, future studies may design and evaluate active learning workflows in which domain experts are asked to annotate only a few cell detections or HTR corrections, and then these corrections can be propagated throughout the entire processing pipeline. This annotation strategy could iteratively refine both handwritten text recognition models and table extraction components under a few-shot learning paradigm, enabling automation performance gains in overall system performance.
Finally, a collaborative KG curation platform can combine automated extraction with expert validation, authority file grounding, and community-contributed schema extensions—turning our current prototype into a production-ready hybrid intelligence system for historical research.

There are also a few limitations and technical extensions of the current pipeline. First, we did not measure the CER for HTR on the given material, as this would require fine-grained, table-aligned transcriptions that are not available and would require substantial new annotation, so a dedicated HTR evaluation protocol for tabular text is left to future research. Second, even the best-performing configuration (Approach-2) preserved cell-level provenance for only 23.19\% of properties because IE was run on concatenated row text and literals were then reverse-mapped to cells; a natural extension is to switch to cell-level extraction where appropriate, which would increase cell-level provenance coverage, although there is trade-off with row-level text context. Third, our KG construction is implemented as a custom Python mapping; implementing it in declarative mapping languages such as RML or YARRRML would make the approach more generic and easier to adapt to new schemas. Additionally, more efficient provenance representations (e.g., using RDF-1.2/n‑Quads-style implementation) could reduce the current triple duplication caused by named graphs. Finally, we did not ground entities in external vocabularies or mint entities with persistent URIs.

\section{Conclusion}
This work introduces a modular, provenance-aware pipeline that bridges document layout analysis, table reconstruction, HTR, and schema-guided KG construction—towards transforming handwritten historical tables into semantically rich, traceable Linked Data. To the best of our knowledge, only \cite{sun2025docs2kg} have attempted KG construction from tabular document images with provenance preservation; however, their approach lacks image-level traceability, making it impossible to inspect or correct KGs when underlying table reconstruction fails or needs verification.

Our evaluation shows modularity is essential for three reasons. First, it enables fair method comparison via shared metrics (mAP for cells, TED/TED-struct for structure). Second, it supports precise error attribution, isolating failures to cell detection, TSR, or HTR. Third, it ensures end-to-end transparency, with every triple linked back to exact image spans. Without modularity, image-to-KG pipelines remain opaque black boxes; with it, every failure becomes diagnosable.

In addition, provenance tracing enables inspection and correction, as exemplified by the military career event extraction case study (Fig.~\ref{fig:event}). In low-quality, high-complexity IE scenarios, often end-to-end automation fails, where fine-grained provenance from image spans to extracted information is essential for scholarly intervention of sensitive historical data. Using a hybrid-AI approach: provenance-rich candidates from machines, grounded validation from experts, our solution offers a viable path to reliable knowledge graphs from challenging archival sources.

While automation performance improvements remain valuable, our core contribution prioritises verifiability and intervention over perfect automation—enabling historians to leverage AI assistance while maintaining scholarly control over interpretation. Future work will measure the effectiveness of the approach from a human-AI collaboration perspective through empirical studies, incorporating an active learning workflow for performance optimisation, and building collaborative KG curation platforms. 

\clearpage
\appendix
\section{Table Reconstruction Evaluation Metrics}
\label{app:tsr_illustration}

\subsection{mean Average Precision}
\begin{figure}[h!]
    \centering
    \includegraphics[width=.9\linewidth]{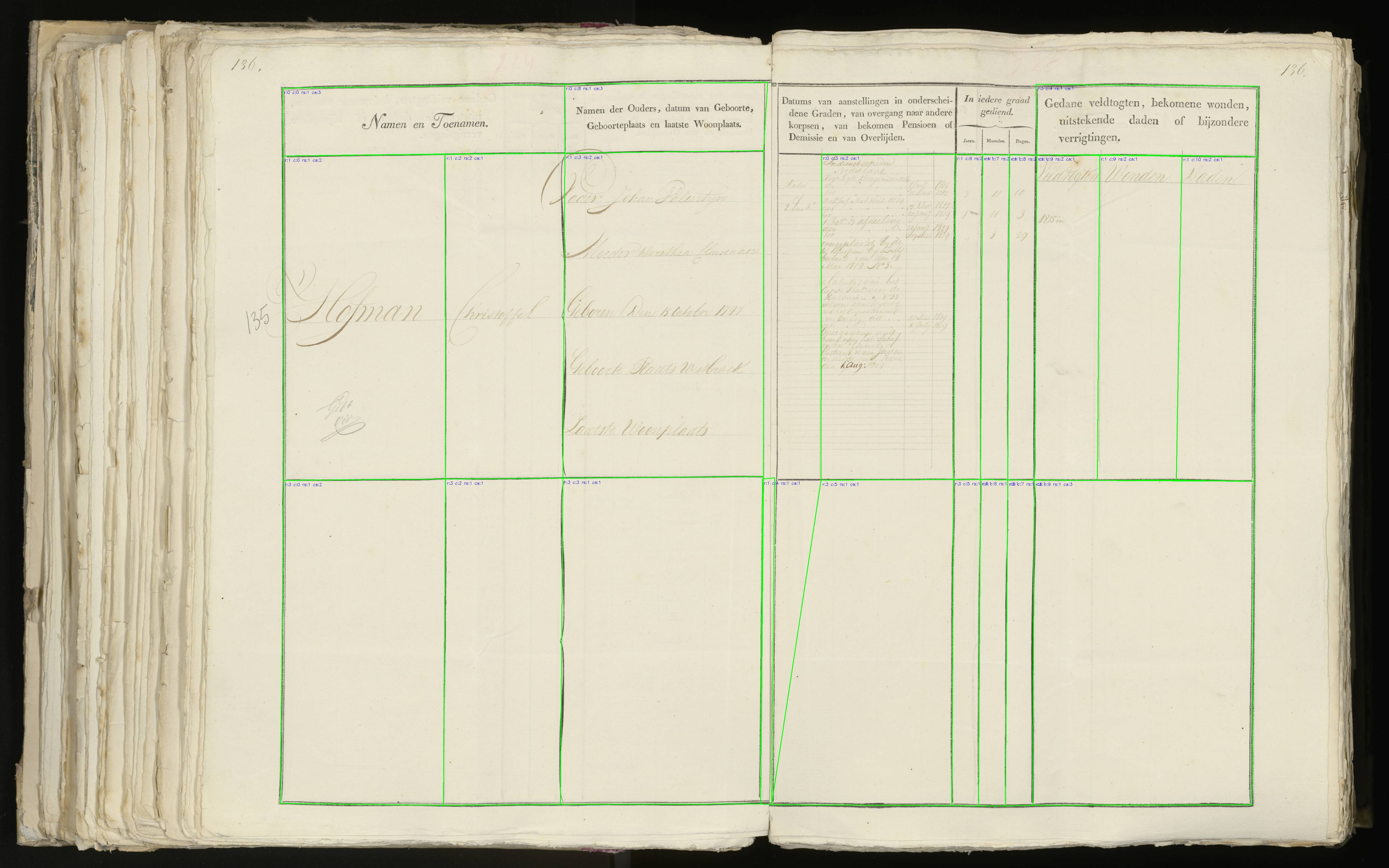}
    \caption{Predicted Cell Bounding Box}
    \label{fig:pred_cell}
\end{figure}

\begin{figure}[h!]
    \centering
    \includegraphics[width=.9\linewidth]{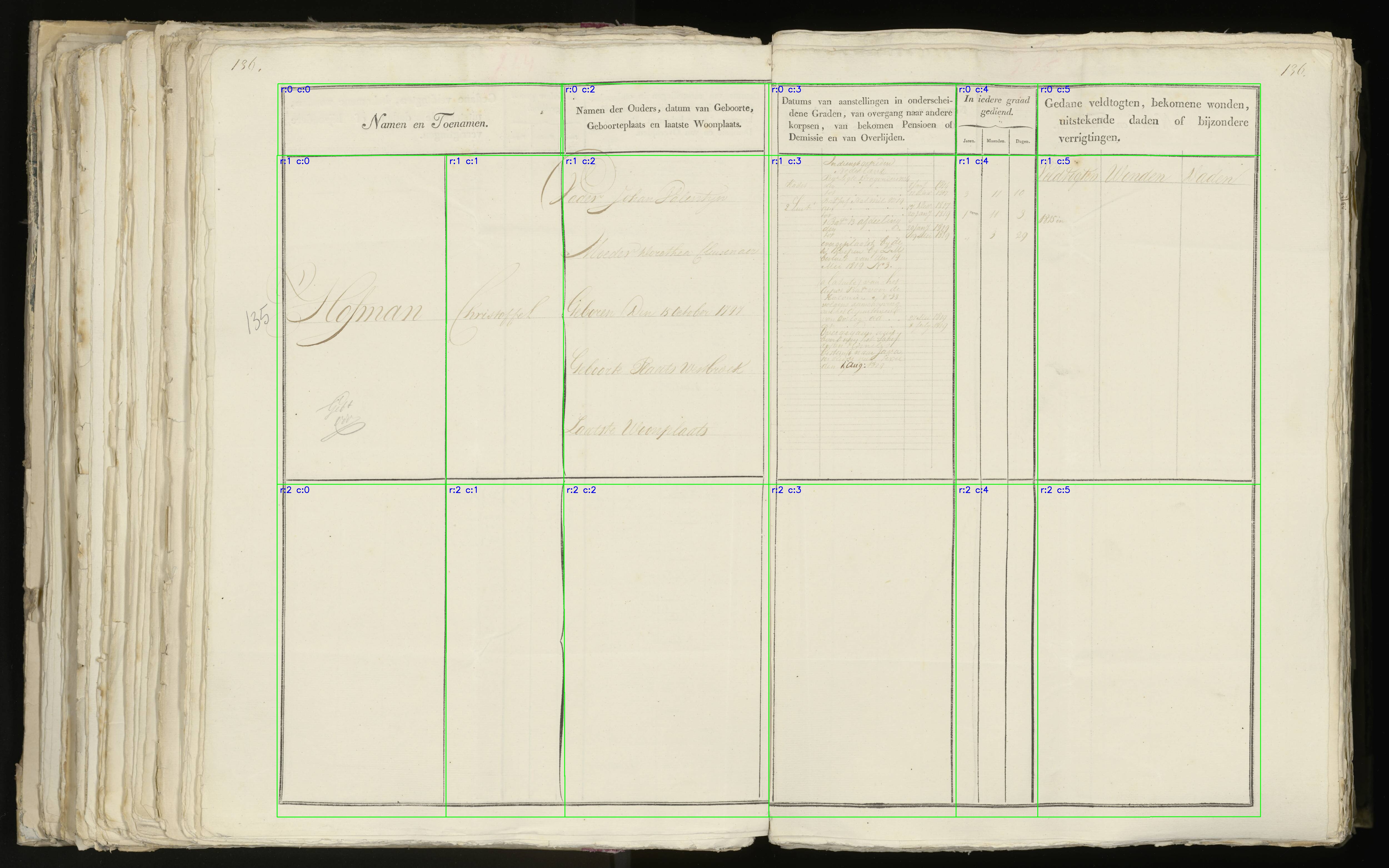}
    \caption{Ground truth Cell Bounding Box}
    \label{fig:gt_cell}
\end{figure}

Figure~\ref{fig:pred_cell} and Figure~\ref{fig:gt_cell} shows predicted and ground truth cell bounding box foe image NL-HaNA\_2.10.50\_45\_0143. From an overview the performance of the automated bounding box looks close enough to the ground truth one, we its calculated maP score is 0.6536.

mean Average Precision (mAP)~\cite{everingham2010pascal} summarizes precision-recall performance across different IoU thresholds. Given predicted and ground-truth cell bounding boxes, we use the Hungarian algorithm~\cite{kuhn2005Hungarian} to optimally align predictions to ground-truth cells by maximizing total IoU, ensuring no double-matching. Predictions with IoU above each threshold are counted as true positives (TP), unmatched predictions as false positives (FP), and unmatched ground-truth cells contribute to false negatives (FN). 

We calculate across continuous IoU thresholds from 0.1 to 1 (step 0.1), computing precision = TP/(TP+FP) and recall = TP/(TP+FN) at each threshold. The mean average precision is calculated as the area under the precision-recall curve across these thresholds using trapezoidal integration.

\subsection{Tree Edit Distance Variants}

\begin{figure}[h!]
    \centering
    \includegraphics[width=.9\linewidth]{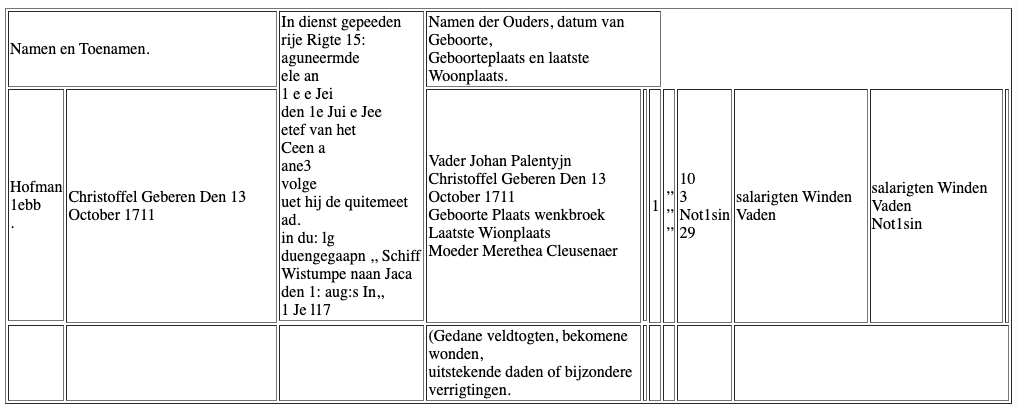}
    \caption{Predicted HTML table}
    \label{fig:pred_html}
\end{figure}

\begin{figure}[h!]
    \centering
    \includegraphics[width=.9\linewidth]{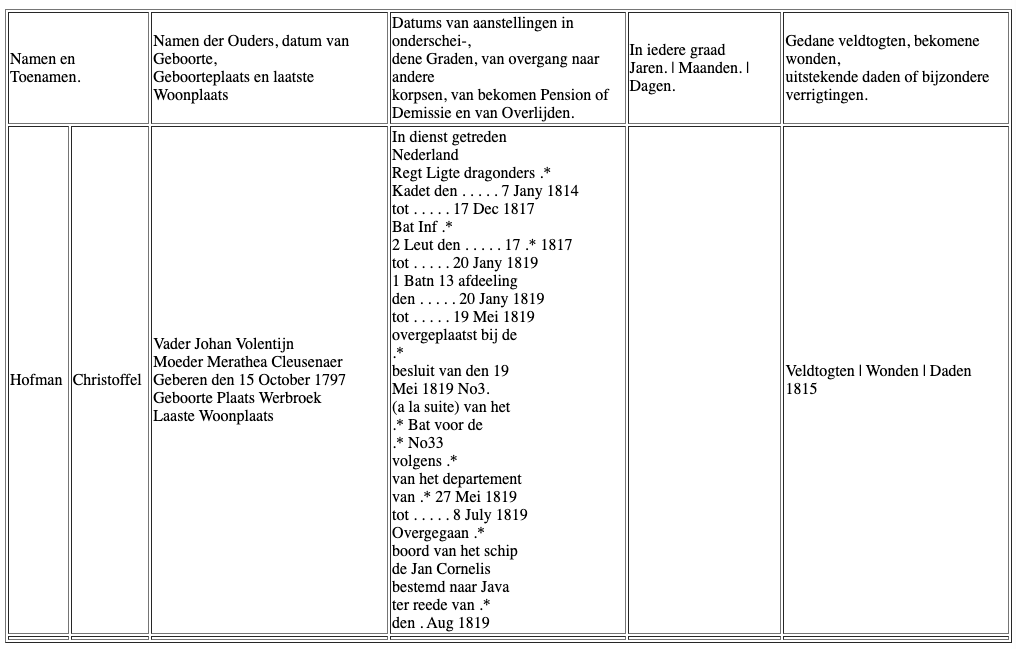}
    \caption{Ground truth HTML table}
    \label{fig:gt_html}
\end{figure}

We implemented Tree Edit Distance (TED) and Tree Edit Distance-struct (TED-strcut) as it is implemented in \cite{zhou2024enhancing}. Figure~\ref{fig:pred_html} and Figure~\ref{fig:gt_html} shows the system generated HTML table and ground truth HTML table for the same image. There corresponding HTML file can be found here. The calculated TEDS and TED-struct score is 0.681 and 0.6949 respectively. 

The TED score is calculated by first parsing tables into hierarchical tree structures, where <table> nodes contain <tr> (row) children, each with <td> (cell) leaf nodes storing colspan, rowspan, and content attributes. The APTED algorithm\furl{APTED library}{https://pypi.org/project/apted/} then computes the minimum edit operations (insert/delete/rename) to transform the predicted tree into the ground-truth tree. So, the TED final score is: 
\begin{equation*}
TED\text{=}1 - \frac{\text{total distance}}{\max\bigl(\lvert node_{\text{pred}} \rvert,\; \lvert node_{\text{true}} \rvert\bigr)}
\end{equation*}

\noindent TED-struct (structure-only) compares tree topology and cell attributes:
\begin{itemize}
    \item Rename cost: 1.0 if tags (td/tr/table) or spans (colspan, rowspan) differ; 0.0 if identical
\end{itemize}

\noindent TED extends this with text content similarity in <td> nodes:
\begin{itemize}
    \item Cell rename cost: Combines structural + content mismatch (0.0 → 1.0 scale)
    \item Tokenization: Each cell is traversed depth-first, extracting tags (<td>, </td>) and text characters
\end{itemize}
\begin{equation*}
    \text{content distance}\text{=}\frac{\text{levenshtein distance}(cell_{\text{pred}},\; cell_{\text{true}})}{\max\bigl(\lvert cell_{\text{pred}} \rvert,\; \lvert cell_{\text{true}} \rvert\bigr)}
\end{equation*}
TED-struct 0.6949 indicates moderate topological fidelity despite mAP=0.325, as minor bounding box errors still preserve row/column hierarchy critical for downstream event extraction. However, the TED score (0.681) drops slightly as it also considers the text content mismatch.

\section{Information Extraction Evaluation Metrics}
\label{app:ie_illustration}
For the evaluation of the Information Extraction component, we compute Precision, Recall, and F1-score for ~\ref{lst:pred_json} by comparing it against ~\ref{lst:gt_json}.

We first apply the Hungarian algorithm \cite{kuhn2005Hungarian} to obtain an optimal one-to-one alignment between the predicted and gold-standard person entities. For each aligned entity pair, corresponding property values are compared using a string similarity metric; values with a similarity score $\geq 0.6$ are treated as correct, thereby contributing one unit to the true positive count. Both literal properties and named-entity properties (including their nested attributes) are uniformly incorporated into the evaluation.

Using this procedure, we obtain a precision of 0.6667, a recall of 0.5714, and an F1-score of 0.6154. The predicted JSON contains four candidate person entities, two of which are empty; however, only the first entity matches the single ground-truth person entity. At the property level, 4 out of the 6 predicted property values are correct and 7 expected property in ground truth, resulting in true positive=4, all predicted=6 and all expected=7.

\begin{lstlisting}[caption={Predicted JSON file for NL-HaNA\_2.10.50\_45\_0143 constructed from predicted HTML},label={lst:pred_json}]
{
  "persons": [
    {
      "name": {
        "basesurname": {
          "value": "Geberen",
          "row": 2,
          "cell": null,
          "original_spans": null
        },
        "firstnames": {
          "value": "Christoffel",
          "row": 2,
          "cell": null,
          "original_spans": null
        }
      },
      "date_of_birth": {
        "value": "13-10-1711",
        "row": 2,
        "cell": null,
        "original_spans": null
      },
      "birth_place": {
        "value": "Wenkbroek",
        "row": 2,
        "cell": "cell_12",
        "original_spans": [
          "118:126"
        ]
      },
      "last_residence": {
        "value": "Winden Vaden",
        "row": 2,
        "cell": [
          "cell_13",
          "cell_18"
        ],
        "original_spans": [
          "212:223",
          "236:247"
        ]
      },
      "country_of_nationality": {
        "value": "Netherlands",
        "row": 2,
        "cell": null,
        "original_spans": null
      }
    },
    {},
    {},
    {
      "name": {
        "basesurname": {
          "value": "Geberen",
          "row": 1,
          "cell": null,
          "original_spans": null
        },
        "firstnames": {
          "value": "Christoffel",
          "row": 1,
          "cell": null,
          "original_spans": null
        }
      },
      "date_of_birth": {
        "value": "13-10-1711",
        "row": 1,
        "cell": null,
        "original_spans": null
      },
      "birth_place": {
        "value": "wenkbroek",
        "row": 1,
        "cell": "cell_12",
        "original_spans": [
          "345:353"
        ]
      },
      "last_residence": {
        "value": "Moeder (assuming this is a location)",
        "row": 1,
        "cell": null,
        "original_spans": null
      },
      "country_of_nationality": {
        "value": "Netherlands (based on the birthplace)",
        "row": 1,
        "cell": null,
        "original_spans": null
      }
    }
  ]
}
\end{lstlisting}

\begin{lstlisting}[caption={Ground truth JSON file for NL-HaNA\_2.10.50\_45\_0143},label={lst:gt_json}]
{
  "persons": [
    {
      "name": {
        "basesurname": {
          "value": "Hofman"
        },
        "firstnames": {
          "value": "Christoffel"
        }
      },
      "date_of_birth": {
        "value": "15-10-1797"
      },
      "birth_place": {
        "value": "Werbroek"
      },
      "last_residence": {
        "value": "Not mentioned"
      },
      "country_of_nationality": {
        "value": "Nederland"
      },
      "military_rank": {
        "value": "2 Lieut"
      }
    }
  ]
}
\end{lstlisting}






\bibliographystyle{vancouver}
\bibliography{bibliography}

\end{document}